# A natural-inspired optimization machine based on the annual migration of salmons in nature

Ahmad Mozaffari[1,*], Alireza Fathi[2]

1: Systems Design Engineering Department, University of Waterloo, ON, Canada

2: Department of Mechanical Engineering, Babol University of Technology, Babol, Iran

* Corresponding Author, E-mail: amozaffari10@yahoo.com

## Abstract

Bio inspiration is a branch of artificial simulation science that shows pervasive contributions to variety of engineering fields such as automate pattern recognition, systematic fault detection and applied optimization. In this paper, a new metaheuristic optimizing algorithm that is the simulation of "The Great Salmon Run" (TGSR) is developed. The obtained results imply on the acceptable performance of implemented method in optimization of complex non-convex, multi-dimensional and multi-modal problems. To prove the superiority of TGSR in both robustness and quality, it is also compared with most of the well-known proposed optimizing techniques such as Simulated Annealing (SA), Parallel Migrating Genetic Algorithm (PMGA), Differential Evolutionary Algorithm (DEA), Particle Swarm Optimization (PSO), Bee Algorithm (BA), Artificial Bee Colony (ABC), Firefly Algorithm (FA) and Cuckoo Search (CS). The obtained results confirm the acceptable performance of the proposed method in both robustness and quality for different bench-mark optimizing problems and also prove the author's claim.

## I. INTRODUCTION

Metaheuristic algorithms are population based methods that work with a set of feasible solutions and try to improve them gradually. These algorithms can be divided into two main categories: Evolutionary Algorithms (EA) which attempt to simulate the phenomenon of natural evolution and swarm-intelligence-based algorithms. There are many different variants of evolutionary algorithms. The common ideas behind these techniques are the same: defining a population of individuals and survival of the fittest according to the theory of evolution. Another branch of population-based algorithms which is known as swarm intelligence focuses on the collective behavior of self-organized systems in order to develop metaheuristics procedures. The interactive behavior between individuals with one another or with their environment contributes to the collective intelligence of the society and often leads to convergence of global behavior. There is a wide variety of swarm-based algorithms which mimic the natural behavior of insects and animals such as ants, fishes, birds, bees, fireflies, penguins, frogs and many others [15, 16]. Particle swarm optimization algorithm (PSO) [4], Mutable Smart Bee Algorithm (MSBA) [17], Bee Algorithm (BA) [5], Artificial Bee Colony (ABC) [3], Firefly Algorithm (FA) [9] and Cuckoo Search (CS) [10] are some of the most well-known swarm base algorithms.

In this paper, a new algorithm called TGSR [11-14] is proposed that is inspired by natural migration of salmons and the dangers which are laid in their migration. In the next section, this phenomenon will be described briefly. Thereafter, all steps for implementing a simulated form of this major natural event will be scrutinized. Then, the features and controlling parameters of the proposed inspired algorithm will be described precisely. At last, it will be compared with cited optimization techniques to elaborate on the strength and weakness of the proposed technique.

## II. THE GREAT SALMON RUN IN NATURE

The salmon run phenomena is one of the great annual natural events happening in the North America where millions of salmons migrate through mountain streams for spawning. Since these creatures provide one of the major food sources for living organisms, their passage upstream is fraught with some grave dangers. Among them, hungry Grizzly bears, human fishers and waterfalls are most crucial dangers that they should face. The hungry Grizzly bears congregate in forested valleys where they forage for whatever food source they can find. However, they can hardly find food and also they are in danger by hungry wolves. Salmons are the most important food sources for these hungry bears. Bears communicate with each other to find a passage with higher amount of chubby salmons. In fact, they follow the swarm intelligence rules for hunting salmons with higher qualities. Humans are one of the other major hunters of salmons. These fishers often congregate in Alaska where there is an appropriate condition for hunting plenty of salmons. Humans often mimic some diverse heuristic methodologies to find a region that possesses salmons with higher quality and quantity. They employ scout ships for investigating the complete passage region. The rest of the fishers are integrated in areas with higher salmon’s intensity. *Fig. 1* illustrates a scheme of great salmon run phenomena.

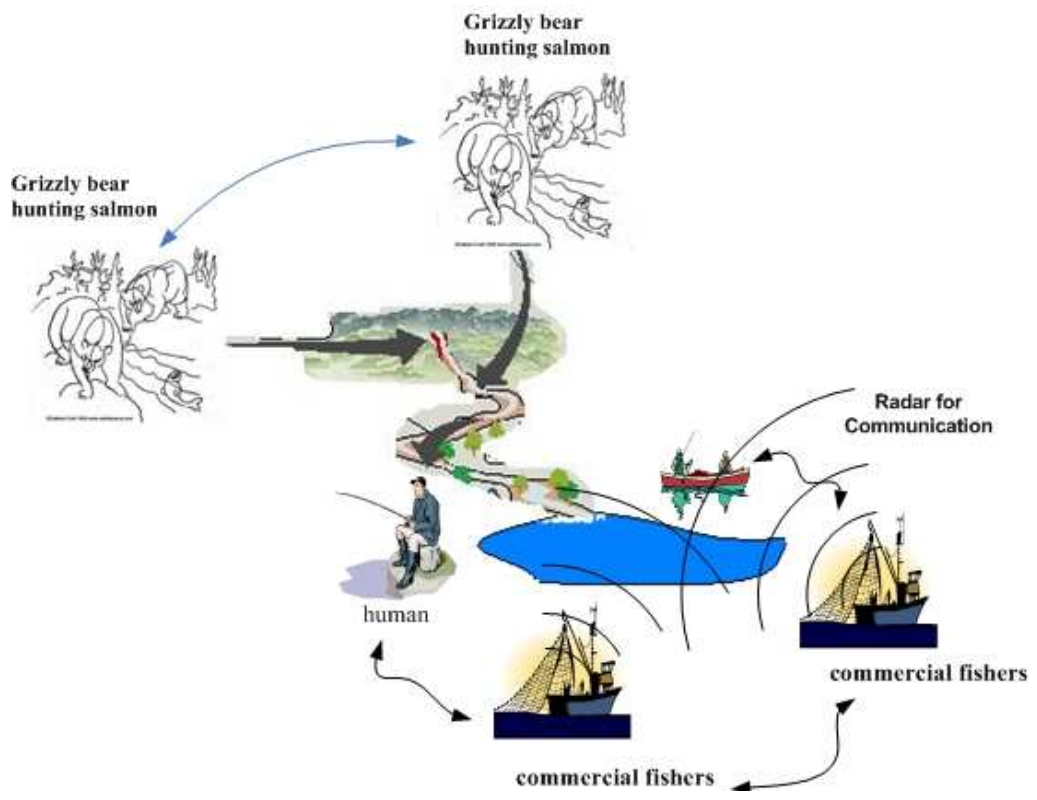

Figure 1. A scheme of great salmon run : grizzly bear and commercial fishers

The experiments prove that the humans' hunting technique is really effective since they first separate into some subparts to investigate the whole area and then integrate in intense areas. There are still many other elements that menace salmon's life during their migration. Waterfalls are another major danger since they can be found in all pathways. At the beginning of the salmon's migration, they divide into subgroups using their instinct and some stochastic interprets. Each of these groups follows different pathways to their destination. Some of them choose forested passages which are full of different dangerous hunters such as Grizzly bears and salmon sharks while others move towards oceans, lakes and ponds. Released annual official reports about the incremental quality and quantity of hunted salmons, suggests a high potential of this great natural phenomenon to be simulated in the form of a parallel evolutionary strategy. Here, this concept has been utilized to create a novel intelligent optimization algorithm.

In the next section, all of the abovementioned stages that occur during the natural salmon's migration will be applied in the simulation of TGSR algorithm. Thereafter, the resulted algorithm will be compared with other intelligent optimization techniques.

## III. THE GREAT SALMON RUN (TGSR) ALGORITHM

The proposed TGSR contains two independent intelligent evolutionary operators [18]. Each of these two operators delegates an independent salmon's migration pathway. The first one belongs to the salmons that move through forested regions and mountain's canyons. The other one belongs to the salmons which are passed through oceans, lakes and ponds. As it was mentioned in the former section, salmons choose their passage based on their instinct and without a meaningful inference. However, the experimental evidence shows that these creatures prefer to cross from ponds and lakes rather than canyons and forested passages [7]. Each of these ways is incorporated with their own natural menaces. Commercial fishers are concentrating on the ponds and ocean pathways while Grizzly bears hunt the salmons that pass through mountain's canyons and forested regions [8]. Each of these two main hunters utilizes different techniques for hunting salmons with higher qualities. TGSR utilized all above steps to handle an optimization problem. Here, the main steps of the algorithm are given.

### *A. Initialization*

In the proposed algorithm, each potential solution delegates the salmon intensity in a region (amount of salmons in a sub group). In other words, region with higher salmon intensity yields a solution with higher fitness. The solutions are initialized stochastically spanning to the passage dominance (between lower bound and upper bounds). (1) represents a procedure which is used to initialize random solutions with respect to the solution space.

$$Initial\ solution = lb + rand * (ub - lb) \quad (1)$$

where lb and ub are the lower and upper bounds, respectively and rand is a random number spanning to 0 and 1 with a uniform distribution.

After initializing the solutions, optimization procedure is started. At the beginning of the optimization process, all of these initialized solutions (salmons sub groups) are prepared for their migration (iterative movements). It is obvious that each iteration cycle is equivalent to a natural migration phenomenon.

### *B. Choosing Pathways for Migration*

Before migration, salmons choose their pathway based on their instinct. This suggests a stochastic shuffling control parameter for thrusting the salmon groups (initial solutions) in both pathways (evolutionary operators). (2) formulates a mathematical form of this process.

$$Solution's\ Sharing: \begin{cases} N_{P_1} = [\,\mu * P_s\,] \\ N_{P_2} = P_s - N_{P_1} \end{cases} \quad (2)$$

where $N_{P_1}$ is the number of salmon groups passing through ocean and ponds, $N_{P_2}$ is the number of salmon group passing through forested regions and mountain canyons, $P_s$ is the number of all salmon groups which participate in the migration and μ is a sharing factor that represents the salmon's instinct.. As seen, the proposed formulation is a strategy for shuffling the solutions stochastically. The results will confirm the effectiveness of this shuffling in the diversification of the solutions.

After exerting the sharing process, these subgroups are entered in their pathway (evolutionary operator). They face different dangers while crossing these pathways. In the following, the details of the passage traversing are given.

### *C. Crossing Lakes and Ponds*

In the first operator, the human hunting is simulated. Humans hire scout ships to investigate the passage dominance (solution space). These scout ships applies some arithmetic graphical search (an intelligent diversification methodology) to explore the passage as best as they can. This exploration has been mathematically modeled in (3).

$$\begin{cases} X_N = X_F + \delta\big(t, (ub - X_F)\big) \\ \quad\quad or \\ X_N = X_F + \delta\big(t, (X_F - lb)\big) \end{cases} \quad (3)$$

where $t$ represents the current iteration number, $X_N$ represents a new detected region (new solution) and $X_F$ shows the former region of the scout ship (former solution). $\delta(x, y)$ is calculated using (4).

$$\delta(x, y) = y * rand * (1 - \frac{x}{T})^b \quad (4)$$

where $T$ is the number of the maximum iteration, $b$ is a random number larger than 1 and $rand$ is a random number spanning to 0 and 1 with a uniform distribution.

The remaining ships (known as commercial fishers) communicate with both scout ships and other active commercial fishers to find better areas (with higher salmon intensity) for hunting salmons. Then they congregate in the areas with higher intensity of salmons. Fisher groups often consist of two main hunter ships and one recruited ship. First, the main hunters find regions with an acceptable salmon intensity (solution fitness). After that, they inform the recruited agent to exploit nearby regions to find more intense areas (solution with higher fitness). This exploitation has been mathematically modeled in (5).

$$X_R = \beta * (X_{M1} - X_{M2}) + X_{M1} \quad (5)$$

where β is a random number spanning to 0 and 1 with uniform distribution, XR represents the new detected solution by the recruited agent, $X_{M1}$ is the solution obtained by the first main hunter and $X_{M2}$ is the solution obtained by the second one.

### *D. Crossing Mountain Canyons and Forested Regions*

The second operator simulates the Grizzly bears hunting methodology. Similar to other animals, Grizzly bears communicate with each other to find a region with higher salmon intensity. Their hunting method is really simple. They always inform each other if they find an acceptable region. Then, the entire Grizzly bear groups approach the best region and search nearby areas. If they find an area with higher salmon intensity, they inform other bears. Otherwise, they leave the region and continue the local search. One of the main disadvantages of the bears hunting procedure is the lack of an independent diverse exploration. Bears hunting methodology is mathematically expressed in (6).

$$X_B = cos\ (\varphi) * (B_R - L_R) + B_R \quad (6)$$

where $X_B$ represents a new detected region, $B_R$ is the best reported region by the hunting team, $L_R$ is the current region for which the bears have decided to perform a local exploitation and φ is an arbitrary angle spanning to 0 and 360 degrees. cos (φ) directs the bears to their destination. *Fig. 2* shows the schematic of Grizzly bear's movements around the best solution. It is obvious that these animals perform an exploitation search with different radii and angle of attacks.

### *E. Regrouping for Spawning*

At the end of the migration, the survived salmons congregate in their destination for spawning. In TGSR, this natural event is simulated through a collection container. After salmons pass through their pathways (operator's performance), the salmon subgroups (solutions) are collected in a unique container. In other words, the solutions are extracted from both operators and make a unique population. At this state, the algorithm has reached the end of the first iteration.

The change in climate and urge for spawning are two main motivations which force the remaining salmons to begin another migration. Continuity of these permanent migrations turns the TGSR to a powerful iterative optimization algorithm. *Fig. 3* illustrates the flowchart of the proposed TGSR algorithm.

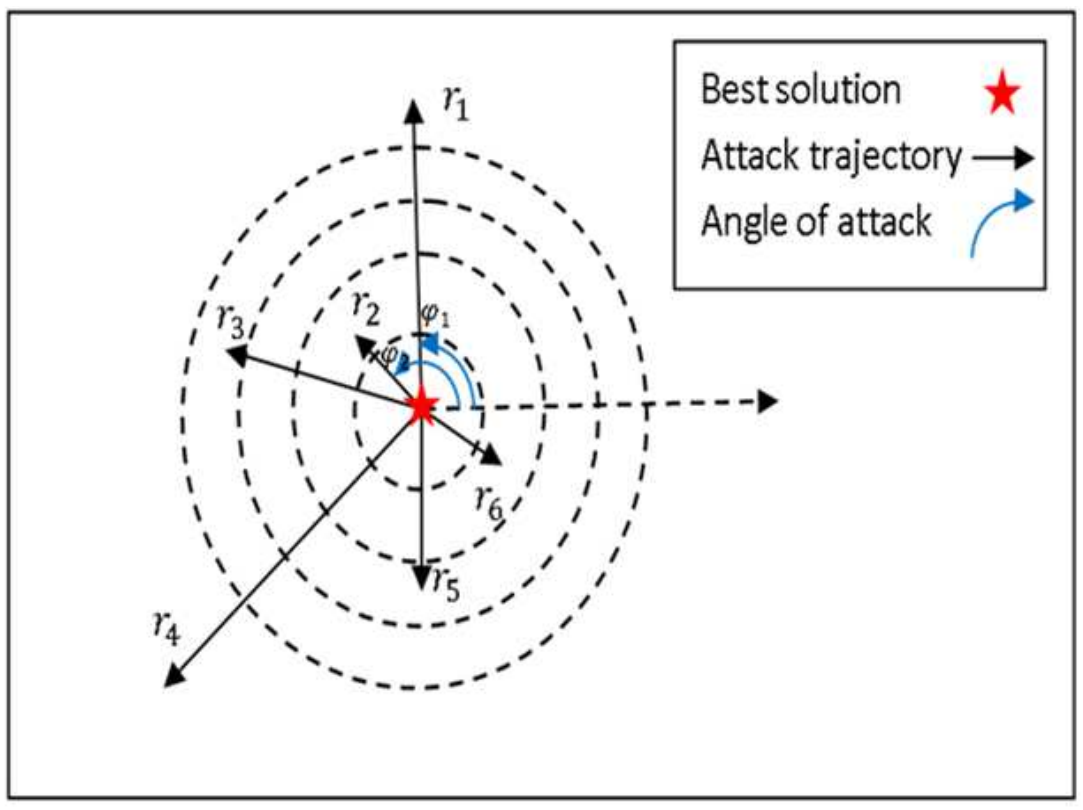

Figure 2. Schematic of Grizzly bear's movemen

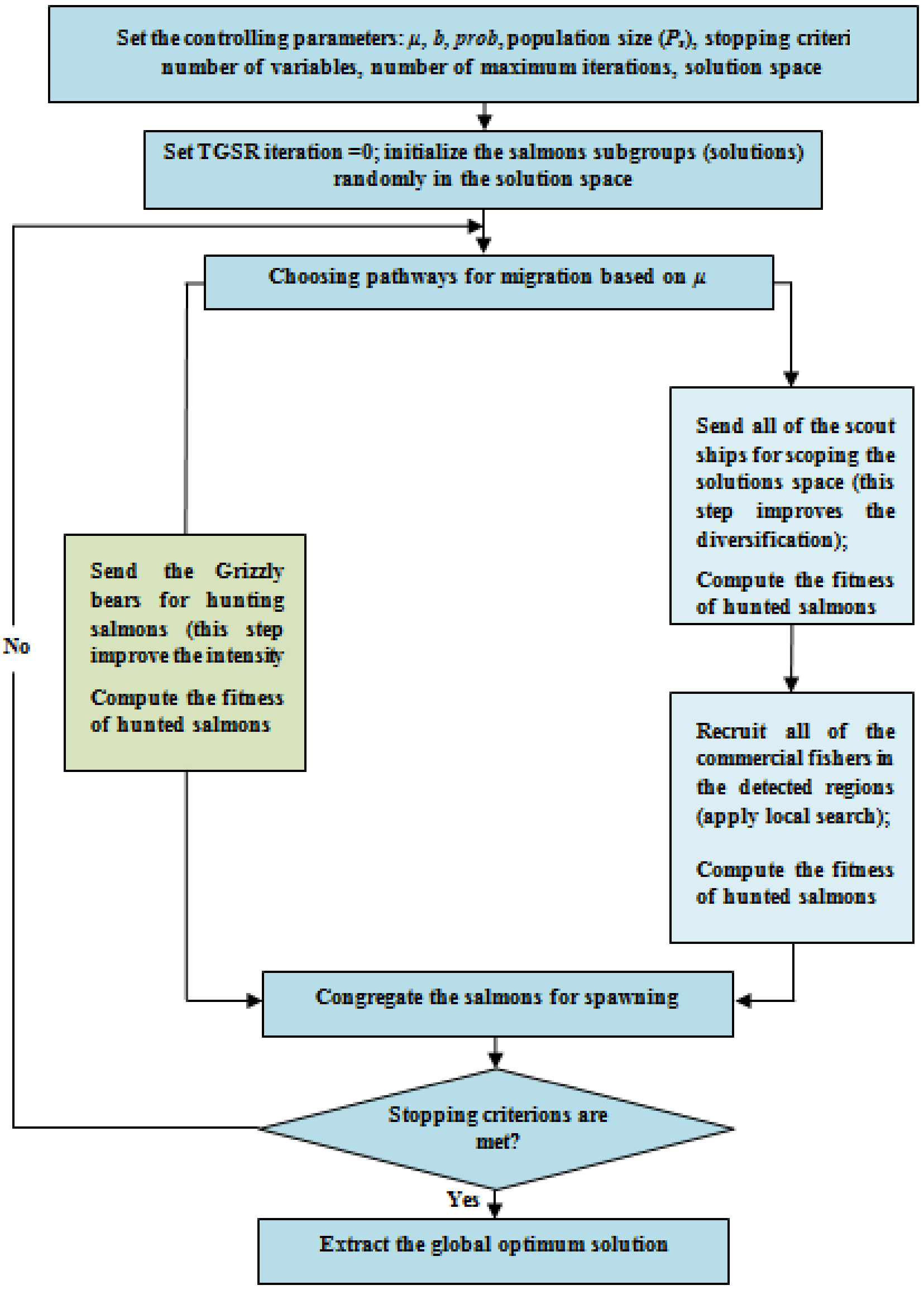

Figure 3. The flowchart of the TGSR algorithm

In the following section, different numerical benchmark optimization problems are applied to elaborate on the high robustness and acceptable quality of the TGSR algorithm.

## IV. Results and Discussion

### A. *Definition of the Control Parameters and Comparison Metrics*

For validating the performance of the proposed optimization algorithm, the obtained results have been compared with most of the well-known optimization techniques. The comparison metrics are measure of the quality and robustness of the algorithm. For all benchmark optimization problems, the maximum iteration number is the stopping criterion. Table 1 indicates the control parameters which are considered in this case study.

In the next section, the performance of TGSR will be compared to other optimization algorithms.

### B. *TGSR for Benchmark Probleems*

In this section, the performance of TGSR will be tested for optimizing some well-known unconstraint problems (functions). These functions are extracted from Belkadi et al. [2]. All algorithms were executed in 30 different runs with different initial random seeds. Table 2 compares the quality (mean) and standard deviation (robustness) of the tested algorithms in optimization problems with 30 independent dimensions.

It seems that TGSR is an optimization method with high robustness and quality in optimizing the stochastic test functions with high dimensionality. It is also evident that the proposed algorithm dominates all other techniques in both robustness and quality. In the first case, DEA, CS and TGSR reached the global solution for the Schaffer function. It can also be seen that BA and SA show weaker results comparing to the other optimization algorithms. The robustness and quality of ABC and PSO are acceptable in this case. The results show that FA and SA possessed 95% and 90% success respectively which are acceptable in this case. Again, for Sphere function, all of the optimization algorithms reach the local optimum condition. However, it can be seen that TGSR and PMGA reach better optimum solutions compared to the other optimization techniques. For Griewank function, PMGA, DEA and TGSR find the global optimum solution with 100% success. It shows the high reliability of these techniques in optimizing unimodal problems. ABC, PSO and SA find an acceptable local optimum solution and FA and CS show weaker results in both quality and robustness. In optimizing the Rastrigin function; it is only TGSR that is completely successful in finding the global optimum solution. CS shows an acceptable quality in finding a nearly optimum solution. It can also be seen that BA, PSO, ABC and DEA show weak results in optimization of the Rastrigin function. In the last case, TGSR shows an explicit predominance (100% success) in optimization of the Rosenbrock problem versus other algorithms. After above numerical investigations, it can be concluded that the TGSR algorithm is highly reliable in optimizing unconstraint problems.

## V. CONCLUSIONS

In this paper, a novel bio inspired metaheuristic method called TGSR has been proposed that is implemented based on the natural phenomenon of salmon’s migration and the dangers that are laid behind their passages. To confirm the high robustness and efficient performance of proposed method, it has been compared with most of the famous optimization techniques. The obtained results were really promising.

After that the algorithm has been applied for optimal designing of spring elements. The gained results show the high potential of TGSR for dealing with serious engineering problems. In the next work, the effects of controlling parameters on the convergence speed and diversification of proposed algorithm will be verified more closely. Besides, a multi objective form of TGSR will be introduced for optimizing some real life engineering problems.

TABLE I. CONTROL PARAMETERS OF THE BENCHMARK OPTIMIZATION ALGORITHMS

| Algorithm | Parameters | | | |
|---|---|---|---|---|
| **TGSR** | $\mu$= 0.75 | $P_s$= 40 | $iter$=10 | $b$=1.6<br>$WFP$=0.1 |
| **PMGA**<br>[2] | $P_{Crossover}$= 0.8 | $P_s$=100 | $iter$=100 | $P_{Mut}$=0.05<br>$MigNum$=3 |
| **ABC**<br>[3] | $Limit$= 10 | $NP$=200 | $iter$=100 | $Prob$=0.1 |
| **PSO**<br>[4] | $nSW$=100 | $iter$=100 | $IW$=0.72 | c1= 2<br>c2=2 |
| **FA,** Ref.<br>[9] | $FN$=50 | $iter$=100 | $\alpha$=0.5 | $\beta$=0.2<br>$\gamma$=1.0 |
| **CS**<br>[10] | $Nests$=40 | $iter$=100 | $Pa$=0.25 | $\alpha$=0.9 |
| **SA**<br>[1] | $Material$=1 | $iter$=100 | $gaf$=0.95 | |
| **DEA**<br>[6] | $P_s$=50 | $iter$=100 | $F$=1.25 | $CR$=0.3 |
| **BA**<br>[5] | $n$=100<br>nep=25 | $iter$=100<br>nsp=10 | $e$=8<br>ngh= rang/10 | $m$=20 |

TABLE II. PERFORMANCE OF THE OPTIMIZATION ALGORITHMS FOR UNCONSTRAINT PROBLEMS WITH A DIMENSION OF 30

| | | ***Problems (Dimensions=30)*** | | | | |
|---|---|---|---|---|---|---|
| ***Algorithms*** | | ***Schaffer*** | ***Sphere*** | ***Griewank*** | ***Rastrigin*** | ***Rosenbrock*** |
| **ABC** [1] | **Quality** ***(Success %)*** | 11.43E+000 (100%) | 356.06E+000 (100%) | 1.27E-001 (100%) | 989.44E+000 (100%) | 123.72E+000 (75%) |
| | **Robustness** ***(Success %)*** | 44E+000 (100%) | 151.03E+000 (100%) | 14.44 E-002 (100%) | 165.29E+000 (100%) | 209.87E+000 (75%) |
| **PMGA** | **Quality** ***(Success %)*** | 12.63E+000 (100%) | 34.43E+001 (100%) | **0** **(100%)** | 120.34E+000 (90%) | 101.11E+000 (90%) |
| | **Robustness** ***(Success %)*** | 23.55E-019 (100%) | 4.92E+000 (100%) | **0** **(100%)** | 23.05E+000 (90%) | 342.72E+000 (90%) |
| **BA** [9] | **Quality** ***(Success %)*** | 23.52E+000 (66%) | 3425.34E+000 (78%) | 23.84E+000 (100%) | 6823.32E+000 (70%) | 1142.45E+001 (77%) |
| | **Robustness** ***(Success %)*** | 29.76E+000 (66%) | 2131.02E+000 (78%) | 21.96E+000 (100%) | 5644.32E+000 (70%) | 4591.47E+000 (77%) |
| **PSO** [10] | **Quality** ***(Success %)*** | 0.53E-007 (88%) | 3110.12E-008 (74%) | 11.33E+000 (67%) | 1322.11E+000 (80%) | 4545.23E+000 (58%) |
| | **Robustness** ***(Success %)*** | 9.00E-008 (88%) | 2314.65E+000 (74%) | 78.95E+000 (67%) | 121.44E+000 (80%) | 2242.12E+000 (58%) |
| **SA** | **Quality** ***(Success %)*** | 27.53E+000 (90%) | 8554.34E+000 (88%) | 19.88E+000 (%90) | 7752.98E+000 (66%) | 29.65E+000 (63%) |
| | **Robustness** ***(Success %)*** | 18.00E+000 (90%) | 1111.02E+000 (88%) | 12.46E+000 (90%) | 1112.76E-001 (66%) | 16.61E+000 (63%) |
| **DEA** [1] | **Quality** ***(Success %)*** | **0** **(100%)** | 21.21E-004 (68%) | **0** **(100%)** | 235.31E+000 (83%) | 143.67E+000 (100%) |
| | **Robustness** ***(Success %)*** | **0** **(100%)** | 34.00E+000 (68%) | **0** **(100%)** | 27.44E+001 (83%) | 20.23E+001 (100%) |
| **CS** [10] | **Quality** ***(Success %)*** | **0** **(100%)** | 260.78E-001 (90%) | 210.78E+000 (100%) | 10.34E+000 (72%) | 2.12E+002 (100%) |
| | **Robustness** ***(Success %)*** | **0** **(100%)** | 311.13E-001 (90%) | 110.02E+000 (100%) | 4.10E+000 (72%) | 5.06E+001 (100%) |
| **FA** [9] | **Quality** ***(Success %)*** | 6.63E-015 (95%) | 19.57E+000 (80%) | 211.32E+000 (100%) | 1323.39E+000 (90%) | 19.76E+000 (97%) |
| | **Robustness** ***(Success %)*** | 12.31 (95%) | 144.73E+000 (80%) | 145.45E+000 (100%) | 1231.07E+000 (90%) | 25.69E+000 (97%) |
| **TGSR** | **Quality** ***(Success %)*** | **0** **(100%)** | **5.33E-017** **(100%)** | **0** **(100%)** | **1.23E-002** **(97%)** | **1.04E-002** **(100%)** |
| | **Robustness** ***(Success %)*** | **0** **(100%)** | **11.09E-023** **(100%)** | **0** **(100%)** | **0.34E-014** **(97%)** | **2.96E-009** **(100%)** |